\documentclass[sigconf]{acmart}

\AtBeginDocument{%
  }

\copyrightyear{2026}
\acmYear{2026}
\setcopyright{cc}
\setcctype{by}
\acmConference[WWW '26]{Proceedings of the ACM Web Conference 2026}{April 13--17, 2026}{Dubai, United Arab Emirates}
\acmBooktitle{Proceedings of the ACM Web Conference 2026 (WWW '26), April 13--17, 2026, Dubai, United Arab Emirates}
\acmPrice{}
\acmDOI{10.1145/3774904.3792124}
\acmISBN{979-8-4007-2307-0/2026/04}

\usepackage{comment}
\usepackage{amsmath, mathrsfs}
\usepackage{graphicx}
\usepackage{enumitem}
\usepackage{cite}
\usepackage{pgfplots}
\pgfplotsset{compat=1.15}
\usetikzlibrary{arrows}
\usepackage{soul}
\usepackage{array, booktabs, multirow, xcolor, colortbl, caption}
\usepackage{makecell}
\usepackage{pifont}
\usepackage{wrapfig}
\usepackage{subcaption}

\newcommand{\Interleave}{\operatorname{Interleave}}
\newcommand{\taesar}{\textsc{Taesar}\space}

\newcommand{\argmax}[1]{\underset{#1}{\operatorname{arg}\,\operatorname{max}}\;\ }
\DeclareMathOperator{\Regenerate}{Regenerate}

\begin{document}

\title{Generative Data Transformation: From Mixed to Unified Data}

\author{Jiaqing Zhang}
\orcid{0009-0001-1039-9735}
\email{jiaqing.zhang@mail.ustc.edu.cn}
\affiliation{%
  \institution{University of Science and Technology of China, Hefei, China}
  \country{}
}

\author{Mingjia Yin}
\orcid{0009-0005-0853-1089}
\email{mingjia-yin@mail.ustc.edu.cn}
\affiliation{%
  \institution{University of Science and Technology of China, Hefei, China}
  \country{}
}

\author{Hao Wang}
\authornote{Corresponding author.}
\orcid{0000-0001-9921-2078}
\email{wanghao3@ustc.edu.cn}
\affiliation{%
  \institution{University of Science and Technology of China, Hefei, China}
  \country{}
}

\author{Yuxin Tian}
\orcid{0009-0008-6141-6284}
\email{tyx682@mail.ustc.edu.cn}
\affiliation{%
  \institution{University of Science and Technology of China, Hefei, China}
  \country{}
}

\author{Yuyang Ye}
\orcid{0000-0002-1513-7814}
\email{yeyuyang@mail.ustc.edu.cn}
\affiliation{%
  \institution{University of Science and Technology of China, Hefei, China}
  \country{}
}

\author{Yawen Li}
\orcid{0000-0003-2662-3444}
\email{warmly0716@126.com}
\affiliation{%
  \institution{Beijing University of Posts and Telecommunications, Beijing, China}
  \country{}
}

\author{Wei Guo}
\email{guowei67@huawei.com}
\orcid{0000-0001-8616-0221}
\affiliation{%
 \institution{Huawei Noah’s Ark Lab}
 \city{Shenzhen}
 \country{China}
}

\author{Yong Liu}
\email{liu.yong6@huawei.com}
\orcid{0000-0001-9031-9696}
\affiliation{%
 \institution{Huawei Noah’s Ark Lab}
 \city{Shenzhen}
 \country{China}
}

\author{Enhong Chen}
\authornote{Corresponding author.}
\orcid{0000-0002-4835-4102}
\email{cheneh@ustc.edu.cn}
\affiliation{%
  \institution{University of Science and Technology of China, Hefei, China}
  \country{}
}

\renewcommand{\shortauthors}{Jiaqing Zhang et al.}

\begin{abstract}
Recommendation model performance is intrinsically tied to the quality, volume, and relevance of their training data. To address common challenges like data sparsity and cold start, recent researchs have leveraged data from multiple auxiliary domains to enrich information within the target domain. However, inherent domain gaps can degrade the quality of mixed-domain data, leading to negative transfer and diminished model performance. Existing prevailing \emph{model-centric} paradigm -- which relies on complex, customized architectures -- struggles to capture the subtle, non-structural sequence dependencies across domains, leading to poor generalization and high demands on computational resources. To address these shortcomings, we propose \textsc{Taesar}, a \emph{data-centric} framework for \textbf{t}arget-\textbf{a}lign\textbf{e}d \textbf{s}equenti\textbf{a}l \textbf{r}egeneration, which employs a contrastive decoding mechanism to adaptively encode cross-domain context into target-domain sequences. It employs contrastive decoding to encode cross-domain context into target sequences, enabling standard models to learn intricate dependencies without complex fusion architectures. Experiments show \textsc{Taesar} outperforms model-centric solutions and generalizes to various sequential models. By generating enriched datasets, \textsc{Taesar} effectively combines the strengths of data- and model-centric paradigms. The code accompanying this paper is available at~ \textcolor{blue}{\url{{https://github.com/USTC-StarTeam/Taesar}}}.

\end{abstract}



% %% A "teaser" image appears between the author and affiliation
% %% information and the body of the document, and typically spans the
% %% page.
% \begin{teaserfigure}
%   \includegraphics[width=\textwidth]{sampleteaser}
%   \caption{Seattle Mariners at Spring Training, 2010.}
%   \Description{Enjoying the baseball game from the third-base
%   seats. Ichiro Suzuki preparing to bat.}
%   \label{fig:teaser}
% \end{teaserfigure}

% \received{20 February 2007}
% \received[revised]{12 March 2009}
% \received[accepted]{5 June 2009}

%%
%% This command processes the author and affiliation and title
%% information and builds the first part of the formatted document.
\maketitle

\begin{figure}[t!] \centering
    \centering
    \includegraphics[width=1.0\columnwidth]{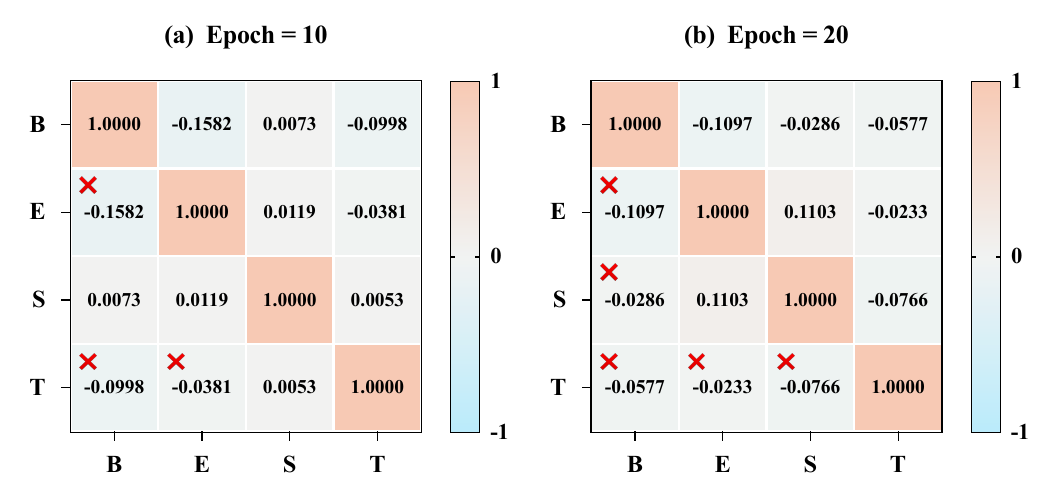}
    \vspace*{-1\baselineskip}
    \caption{Cosine similarity heatmap of gradient directions across four domains in multi-domain sequential recommendation. Naïvely combining data from multiple domains for training induces gradient conflicts and inconsistencies.}
    \vspace*{-1\baselineskip}
    \label{fig:heatmap}
\end{figure}

\section{Introduction}

\vspace{2pt}

\begin{figure*}[t!] \centering
    \centering
    \includegraphics[width=0.99\linewidth]{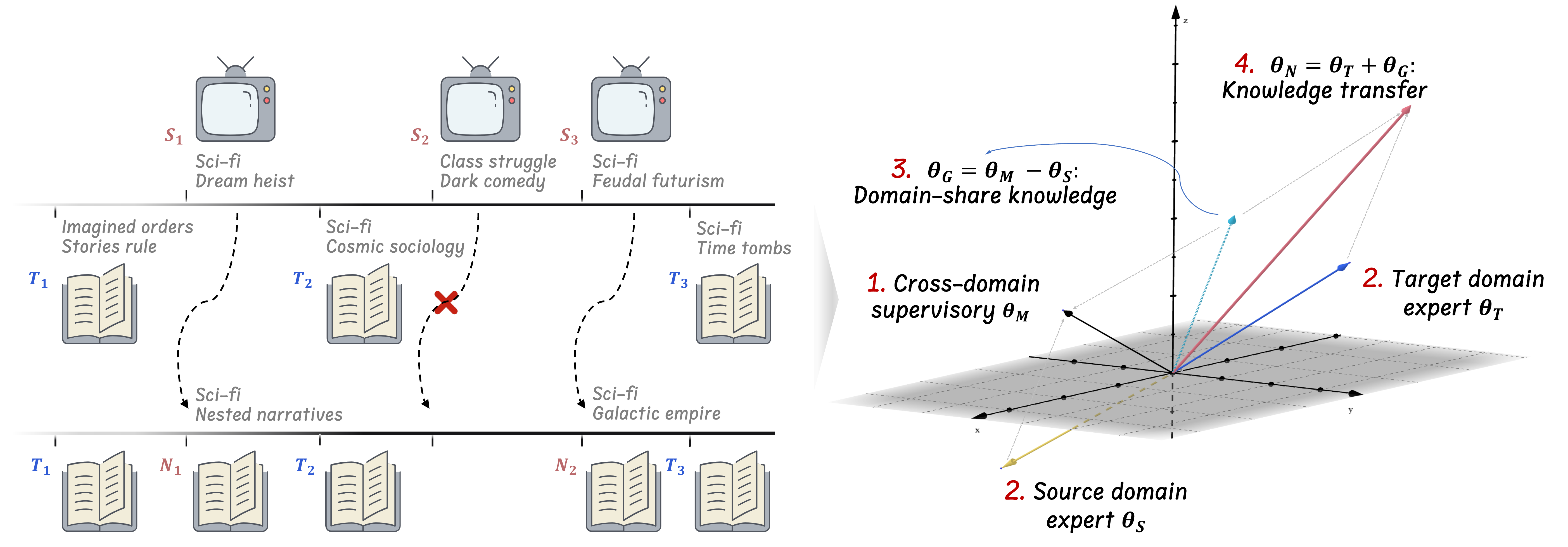}
    \vspace{5pt}
    \caption{Motivation of $\textsc{Taesar}$. We aim to eliminate the domain semantic gap at the source level, prior to model training, thereby transforming cross-domain mixed data into unified target-domain data. \textbf{Left:} Source items with high transferability are mapped to semantically closest target items, while others are discarded. \textbf{Right:} Regenerated data aligns cross-domain behavioral patterns with target-domain relevance, improving model gradients and target-domain performance.}

    \label{fig:motivation}
\end{figure*}

Sequential recommendation seeks to model the temporal dynamics of user preferences based on chronological interaction sequences, thereby facilitating the prediction of users’ subsequent interactions, and enhancing personalized online experiences~\citep{sr-survey1-wu2024survey, sr-survey2-chen2024survey, sr-survey-3-sequential_recommendation_survey1, xie2024breaking, RSS-petrov2024rss, CT4Rec-chong2023ct4rec, shenp}. Although considerable progress has been achieved in this field, real-world user behaviors are frequently dispersed across diverse platforms and domains. This fragmentation introduces two principal challenges: (i) exacerbated data sparsity within individual domains~\citep{BigRec-bao2025bi, METL-kim2023melt, liu2023diffusion, 00-yin2025feature, 01-xie2025breaking, 02-xu2025multi, 03-zhang2025killing, 04-ye2025fuxi, 05-zhou2025multi, 06-wang2025universal, 07-wang2025generative}, which fundamentally limits the modeling capacity of single-domain recommenders, and (ii) significant distributional heterogeneity across domains~\citep{cdsr-survey1-zang2022survey, cdsr-survey2-zhang2025comprehensive, cdsr-survey3-chen2024survey, C2DSR-cao2022contrastive, HGTL-xu2025heterogeneous, SyNCRec-park2024pacer}, wherein domain-specific user interests consequently impede effective knowledge transfer, as shown in Figure~\ref{fig:heatmap}. These challenges collectively prompt a critical research question: How can dispersed cross-domain user activities be effectively harnessed to reliably enhance target-domain recommendation while mitigating the risk of negative transfer~?

\vspace{6pt}

A growing body of research has sought to address this challenge through cross-domain sequential recommendation (CDSR). The dominant paradigm involves jointly learning heterogeneous interaction data from multiple domains within a unified representation space. To mitigate negative transfer, several approaches~\citep{UniSRec-hou2022towards, zhou2025contrastive, park2023cracking, wu2025llm} incorporate auxiliary multimodal information—such as textual or visual content—as bridging signals across domains. For purely ID-based sequences, existing studies typically adopt a \emph{model-centric} paradigm, developing sophisticated cross-domain knowledge transfer and fusion modules, often in conjunction with complex multi-task training strategies~\citep{C2DSR-cao2022contrastive, ABXI-bian2025abxi, SyNCRec-park2024pacer}. Despite their effectiveness, these approaches entail several inherent limitations. On one hand, \textbf{unified multi-domain modeling incurs substantial optimization complexity}: balancing multiple domains within a shared representation space increases model intricacy and training costs, often compromising domain-specific accuracy relative to single-domain counterparts. On the other hand, such methods exhibit \textbf{limited generality and scalability}: their cross-domain fusion modules are typically model-dependent, and multi-task learning setups require extensive hyperparameter tuning, hindering practical deployment.

\vspace{6pt}

To address the limitations of the \emph{model-centric} paradigm, we propose a novel \emph{data-centric} approach for cross-domain sequential recommendation. Instead of designing complex transfer architectures, we focus on refining the data itself by eliminating detrimental inter-domain information and regenerating sequences using only target-domain items (Figure~\ref{fig:motivation}). Consequently, the challenge of multi-domain modeling is transformed into a precise single-domain problem. Building on this concept, we propose $\textsc{Taesar}$ -- target-aligned sequential regeneration. The core insight is that a target-domain model can effectively refine mixed-domain sequences by contrasting its predictions with those from specialized source-domain models, thereby enforcing a closer alignment between the regenerated sequences and target-domain semantics. To operationalize this idea, $\textsc{Taesar}$ first trains a base model on the mixed-domain sequences to capture transferable, global cross-domain patterns. Subsequently, domain-specific expert models are derived by adapting the base model to each domain's individual sequences by \emph{Domain-Specific Prediction} strategy, allowing them to specialize in domain-specific behaviors. Guided by these models, $\textsc{Taesar}$ employs \emph{Adaptative Contrastive Decoding}: for every non-target-domain item within a mixed sequence, it dynamically contrasts the predictions of the corresponding source-domain expert, the target-domain expert, and the base model to determine whether and how to replace the item with a plausible target-domain item. Replacement probabilities are computed using both global and local contrastive scores alongside adaptive weights derived from information theory, yielding refined sequences that preserve the temporal order while containing only target-domain items, consequently enhancing recommendation performance. We highlight four key contributions as follows:

\vspace{-0.5cm}

\begin{itemize}[leftmargin=.15in]
    \item To the best of our knowledge, we present the first \emph{data-centric} paradigm for CDSR, shifting the focus from model-level adaptation to semantic alignment at the data level. By regenerating sequences through contrastive decoding, our approach proactively mitigates negative transfer before model training.

    \vspace{1pt}

    \item We develop a unified generative framework comprising a pretraining and domain-adaptation  module and a adaptive contrastive decoding module, which collectively address two fundamental challenges in $\text{CDSR}$: identifying translatable cross-domain items and preserving inherent behavioral patterns.

    \vspace{1pt}

    \item $\textsc{Taesar}$ generates domain-purified sequences while retaining the original item $\text{ID}$ structure, enabling seamless integration with existing sequential recommenders without any architectural or training modifications—thereby ensuring high practical compatibility and ease of deployment in $\text{CDSR}$ systems.
\end{itemize}

\section{Related works}

\paragraph{\textbf{Cross-Domain Sequential Recommendation (CDSR)}}
CDSR enhances recommendation systems by modeling user behaviors across domains for richer user understanding~\citep{Tri-CDR-ma2024triple}. The mainstream research trajectory has been dominated by \emph{model-centric} paradigm. Early efforts, such as Pi-Net~\citep{pi-Net-ma2019pi} and PSJNet~\citep{PSJNet-sun2021parallel}, employed shared filters and transfer units to exchange information between domains, yet their performance was often constrained by domain similarity and susceptibility to behavioral noise. Subsequent works introduced more advanced transfer mechanisms, including graph-based propagation frameworks (e.g., MIFN~\citep{MIFN-ma2022mixed}, DA-GCN~\citep{DA-GCN-guo2021gcn}) and attention-enhanced representation models (e.g., C2DSR~\citep{C2DSR-cao2022contrastive}, DREAM~\citep{Dream-ye2023dream}, MAN~\citep{MAN-lin2024mixed}) that jointly capture intra- and inter-domain dependencies. To improve generalization and scalability, several studies have explored domain-invariant and text-driven representations. UniCDR~\citep{UniCDR-hou2022towards} and UniSRec~\citep{UniSRec-hou2022towards} leveraged masking and contrastive pretraining to learn transferable user semantics, while RecGURU~\citep{RecGURU-li2022recguru} incorporated adversarial learning to derive domain-agnostic embeddings. More recent approaches have addressed specific challenges, such as mitigating negative transfer via gradient regularization~\citep{SyNCRec-park2024pacer}, aligning semantic spaces for disjoint-user scenarios~\citep{IESRec-liu2023joint}, and enhancing task consistency through invariant LoRA modules~\citep{ABXI-bian2025abxi}. Despite progress, model-centric methods remain plagued by architectural complexity and optimization instability, limiting their adaptability. Therefore, to overcome these hurdles, \textsc{Taesar} aims to shift the paradigm to a \emph{data-centric} view.

\vspace{-5pt}

\paragraph{\textbf{Data Regeneration (DR)}} Machine learning is fundamentally framed as learning a mapping $f: \mathcal{X} \to \mathcal{Y}$, where $\mathcal{X}$ denotes inputs (e.g., interaction history) and $\mathcal{Y}$ denotes targets (e.g., items)~\citep{DR4SR-yin2024dataset}. However, the significant semantic gap between the noisy $\mathcal{X}$ and the concise $\mathcal{Y}$ makes direct optimization difficult. DR addresses this by introducing a latent transition: $\mathcal{X} \to \mathcal{X}' \to \mathcal{Y}$. In this framework, $\mathcal{X}'$ acts as a distilled intermediate representation, simplifying the learning process and enhancing model convergence. This strategy differs fundamentally from existing paradigms such as \textbf{data distillation}~\citep{farzi-sachdeva2023farzi, TD3-zhang2025td3, Distill-CF-sachdeva2022infinite, CGM-wang2023gradient, wu2023tf}, which compresses real data into compact synthetic representations; \textbf{data generation}~\citep{ASReP-liu2021augmenting, METL-kim2023melt, yu2025thought}, which synthesizes new samples from learned distributions; and other \textbf{data augmentation} approaches~\citep{dang2024repeated1, dang2024repeated2, liu2023diffusion, ASReP-liu2021augmenting}, which expand existing datasets through stochastic or semantic transformations. Building on the success of dataset regeneration methods like DR4SR~\citep{DR4SR-yin2024dataset} in capturing transition patterns, we propose to regenerate mixed-domain sequences into target-exclusive formats, a process designed to distill information from multi-domain and boost transferability.

\vspace{-5pt}

\paragraph{\textbf{Contrastive Decoding (CD)}} It enhances language model generation by contrasting expert and amateur outputs to optimize fluency, diversity, and coherence~\citep{CD-li2022contrastive}. Recent advances demonstrate contrastive decoding's versatility across diverse applications: it enhances reasoning in LLMs~\citep{CD-reasoning-o2023contrastive} (despite persistent factual recall challenges), enables universal detoxification without model-specific tuning~\citep{unidetox-huiminunidetox}, improves factual generation through APD's asymptotic probability extrapolation~\citep{APD-chang2024explaining}, and suppresses hallucinations in multimodal models via ECD~\citep{ECD-fieback2025efficient}.  While effective for text generation, CD's potential for data regeneration—especially in sequential recommendation—remains unexplored. Our tri-model CD framework addresses this gap for cross-domain sequential data regeneration. 

\section{Preliminaries}

\paragraph{\textbf{Cross-Domain Sequential Recommendation}}

A domain $\mathcal{D}\strut$ is formally defined as a structured tuple $\mathcal{D} = (\mathcal{U}, \mathcal{V}, \mathcal{E})\strut$, where $\mathcal{U}\strut$ denotes the set of users, $\mathcal{V}\strut$ represents the set of items, and $\mathcal{E}\strut$ captures the set of observed interactions. The interaction set $\mathcal{E}$ comprises a collection of sequences $\{\mathbf{x}_i\}_{i=1}^{|\mathcal{E}|}\strut$, with each sequence $\mathbf{x}_{i} = [x_{ij} \in \mathcal{V}]_{j=1}^{|\mathbf{x}_i|}\strut$ corresponding to an ordered list of items from $\mathcal{V}\strut$, thereby reflecting a user's historical engagement.

In the multi-domain paradigm, we consider $M$ source domains $\{\mathcal{S}_1, \mathcal{S}_2, ..., \mathcal{S}_M\}\strut$ and a singular target domain $\mathcal{T}\strut$. To facilitate domain differentiation, we denote the interaction sequences for a user $u$ as $\mathbf{x}_u^{\mathcal{S}_m} = (x_1^{\mathcal{S}_m}, x_2^{\mathcal{S}_m}, ..., x_t^{\mathcal{S}_m})\strut$ for the $m$-th source domain, and $\mathbf{x}_u^{\mathcal{T}} = (x_1^\mathcal{T}, x_2^\mathcal{T}, ..., x_t^\mathcal{T})\strut$ for the target domain. To enable cross-domain knowledge transfer, we construct a unified, or merged, interaction sequence $\mathbf{x}_{u}^{\mathcal{M}}\strut$ by chronologically interleaving interactions from all source and target domains:
\begin{align}
\mathbf{x}_{u}^{\mathcal{M}} = \Interleave(\mathbf{x}_u^{\mathcal{S}_1}, \mathbf{x}_u^{\mathcal{S}_2}, ..., \mathbf{x}_u^{\mathcal{S}_M}, \mathbf{x}_u^{\mathcal{T}}),
\end{align}
where the $\Interleave(\cdot)$ operator maintains the inherent temporal ordering while consolidating interactions across domains.

Given the observed sequences $\{(\mathbf{x}^{\mathcal{S}_1}, ..., \mathbf{x}^{\mathcal{S}_M}, \mathbf{x}^{\mathcal{T}}, \mathbf{x}^{\mathcal{M}})_u\}\strut$, the cross-domain sequential recommendation task leverages the merged sequence $\mathbf{x}^{\mathcal{M}}\strut$ to improve predictive performance, calculating probabilities based on the integrated interaction history:
\begin{align}
x_{i+1} = \argmax{x_{i+1} \in \mathcal{V}^{\mathcal{T}}} P^{\mathcal{T}}(x_{i+1} | \mathbf{x}_{1:i}^{\mathcal{M}}),
\end{align}
thereby facilitating the vital knowledge transfer from the multiple source domains $\{\mathcal{S}_m\}_{m=1}^M$ to the target domain $\mathcal{T}$.

\paragraph{\textbf{Cross-Domain Data Regeneration}}
Given an input dataset $\mathbf{x}\strut$ where each sequence $\mathbf{x}_u^{\mathcal{M}}\strut$ comprises items originating from multiple source domains and the target domain, data regeneration constructs a novel dataset $\mathbf{x}^{\mathcal{N}}\strut$ consisting exclusively of $\mathcal{T}$-domain items. For a given sequence $\mathbf{x}_{u}^{\mathcal{M}}\strut$, the transformation is defined as:
\begin{align}
\mathbf{x}_{u}^{\mathcal{M}} &= (x_1^{d_1},x_2^{d_2},...,x_{|\mathbf{x}_u^{\mathcal{M}}|}^{d_{|\mathbf{x}_u^{\mathcal{M}}|}}), \quad d_j \in \{\mathcal{S}_1,...,\mathcal{S}_M,\mathcal{T}\}, \\
\mathbf{x}_{u}^{\mathcal{N}} &= (y_1^\mathcal{T}, y_2^\mathcal{T}, ..., y_{|\mathbf{x}_u^{\mathcal{M}}|}^\mathcal{T}) = \Regenerate(\mathbf{x}_u^{\mathcal{M}}), \quad \text{s.t.} \quad \forall ~y_k \in \mathcal{V}^{\mathcal{T}}.
\end{align}

The regeneration operator $\text{Regenerate}(\cdot)$ is implemented as:
\begin{equation}
y_k = \begin{cases}
x_k & \text{if } x_k \in \mathcal{V}^{\mathcal{T}}, \\
f_m(x_k) & \text{if } x_k \in \mathcal{V}^{\mathcal{S}_m}, ~ m=1,...,M,
\end{cases}
\end{equation}
where each function $f_m(\cdot): \mathcal{S}_m \to \mathcal{T} \cup \{\varnothing\}$ either maps (transforms) or discards (replaces with a null item) the item from the $m$-th source domain into the target-domain item space. The composite dataset is subsequently generated as:
\begin{align}
\mathcal{N} = \mathbf{x}^{\mathcal{N}} \cup \mathbf{x}^{\mathcal{T}},
\end{align}
where $\mathbf{x}^{\mathcal{T}}\strut$ retains the original target-domain sequences, while $\mathbf{x}^{\mathcal{N}}\strut$ contains the regenerated target-domain sequences, together constituting an enriched collection of user interaction sequences.

\section{Cross-Domain Sequential Data Regeneration}

\begin{figure*}[t!] \centering
    \centering
    \includegraphics[width=0.85\linewidth]{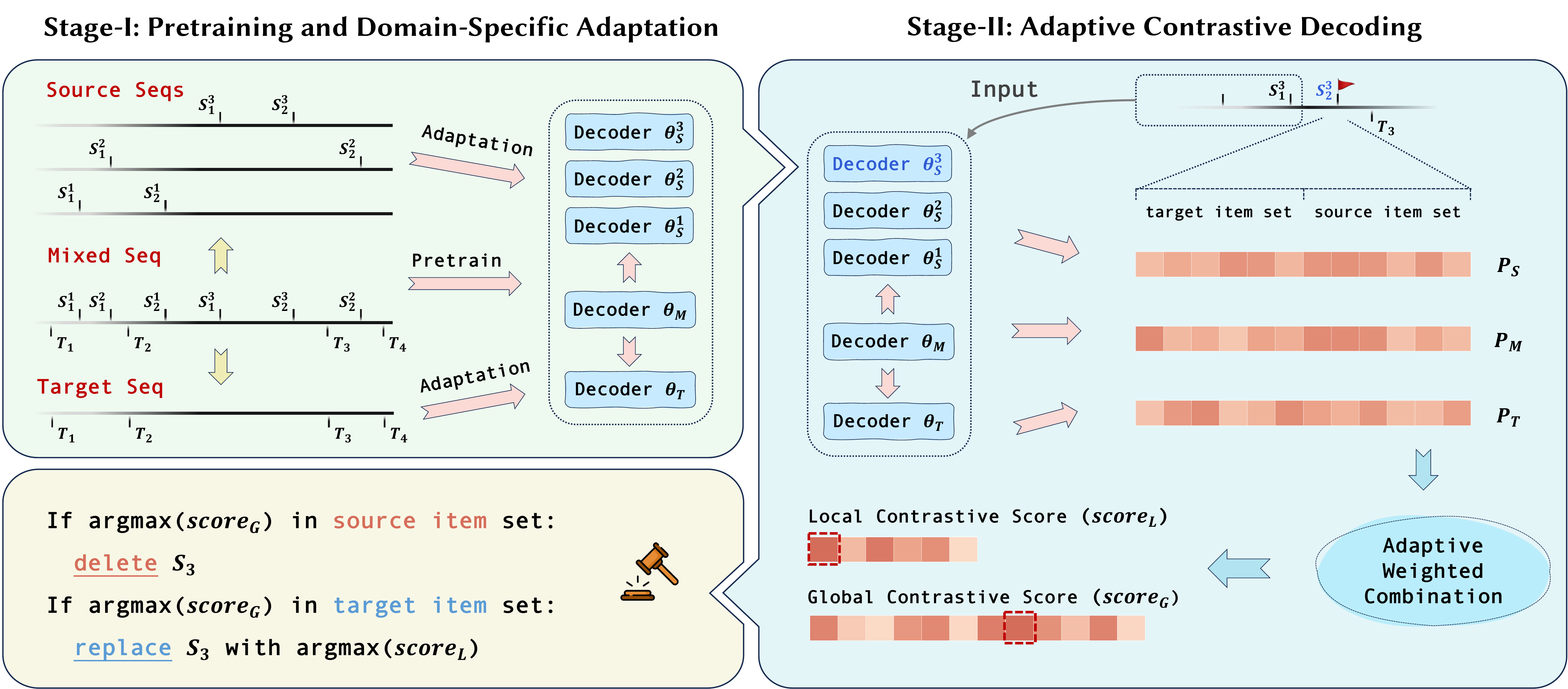}
    % \vspace{3pt}
    \caption{Overview of $\textsc{Taesar}$. (1) In the first stage, \textbf{Tri-model Pretraining}, we construct three views of decoder models: a base model ($\theta_{\mathcal{M}}$), distinct source domain experts ($\theta_{\mathcal{S}}$), and a target domain expert ($\theta_{\mathcal{T}}$). (2) We select the base model, the target domain expert, and the source domain expert corresponding to the item to be transformed to regenerate mixed cross-domain sequences using global and local contrastive decoding among the three models. (3) Based on the two types of contrastive scores, we decide whether to replace a source-domain item with a target-domain item or to discard the source-domain item entirely.}
    \label{fig:overview}
\end{figure*}

In this section, we formally present $\textsc{Taesar}$. As depicted in Figure~\ref{fig:overview}, the framework comprises two principal stages: 1) Pretraining and Domain-Specific Adaptation Stage, designed to preserve cross-domain behavioral patterns and domain-specific semantics; and 2) Adaptive Contrastive Decoding Stage, which focuses on decoupling domain-shared knowledge and eliminating source-domain biases. This dual-mechanism design yields regenerated data that facilitates seamless integration with various downstream recommendation models and ensures strong generalization across subsequent tasks.

\vspace{-5pt}

\subsection{Pretraining and Domain-Specific Adaptation}

Mixed-domain sequences contain items from multiple domains, whose contextual dependencies and behavioral patterns are inherently complex and heterogeneous. Directly applying contrastive decoding to such sequences would make it difficult for the decoder to capture the latent semantic correspondences across domains. Our goal is twofold: (1) to learn the shared cross-domain structural patterns that are transferable across domains, and (2) to develop domain-specific expertise for accurate target-domain predictions. 

% To achieve this, we adopt a two-stage strategy: first, pretrain a base encoder on all domains to capture global patterns; second, adapt domain-specific decoders to specialize in each individual domain while preserving cross-domain knowledge.

\subsubsection{Cross-Domain Encoder Pretraining}

We pretrain a base model $\theta_{\mathcal{M}}$ on mixed-domain sequences to capture shared sequential dynamics and long-term user preferences across all domains. Formally, the pretraining objective is:
\begin{align}
    \theta_{\mathcal{M}}^{*} &\triangleq \arg\min_{\theta} \mathcal{L}_{\mathcal{M}}(\theta~|~\theta^0_{\mathcal{M}}), \\
    \mathcal{L}_{\mathcal{M}}(\theta) &\triangleq \mathbb{E}_{\mathbf{x} \sim \mathbf{x}^{\mathcal{M}}} \left[\sum_{i=1}^{|\mathbf{x}|-1} \ell\left(\Phi_{\theta}(\mathbf{x}_{1:i}), x_{i+1}\right)\right],
\end{align}
where $\Phi_{\theta}$ denotes a Transformer-like decoder, $\ell(\cdot)$ is the cross-entropy loss, and $\mathbf{x}_{1:i}$ is the subsequence of the first $i$ tokens in $\mathbf{x}$. This pretraining stage achieves cross-domain pattern discovery by capturing generalizable behavioral regularities and performs representation alignment by mapping items from all domains into a shared, structured semantic space, providing a structured foundation for the subsequent contrastive decoding stage.

\subsubsection{Domain-Specific Decoder Adaptation}

After obtaining the pretrained encoder $\theta_{\mathcal{M}}^*$, we fine-tune domain-specific decoders for each domain ($\mathcal{S}_1, \mathcal{S}_2, \dots, \mathcal{S}_M, \mathcal{T}$). To preserve cross-domain knowledge while specializing for a target domain $\mathcal{T}$, we employ a \emph{Domain-Specific Prediction (DSP)} strategy: the model receives the full mixed-domain sequence as input but predicts only items belonging to the current domain. Formally, the adaptation objective is:
\begin{align}
    \theta_{\mathcal{T}}^{*} &\triangleq \arg\min_{\theta} \mathcal{L}_{\mathcal{T}}^{\text{DSP}}(\theta~|~\theta_{\mathcal{M}}^{*}), \\
    \mathcal{L}_{\mathcal{T}}^{\text{DSP}}(\theta) &\triangleq \mathbb{E}_{\mathbf{x} \sim \mathbf{x}^{\mathcal{M}}} \left[\sum_{(i,t) \in \mathcal{P}_{\mathcal{T}}(\mathbf{x})} \ell\left(\Phi_{\theta}(\mathbf{x}_{1:i}), x_t\right)\right],
\end{align}
where $\mathcal{P}_{\mathcal{T}}(\mathbf{x})$ contains all valid prediction pairs of target-domain items, that can be formalized as follows:
\begin{align}
    \mathcal{P}_{\mathcal{T}}(\mathbf{x}) = \Big\{(i,t) \mid &~x_i, x_t \in \mathcal{V}^{\mathcal{T}},~ t>i,~ \not\exists j\in(i,t) \text{ s.t. } x_j \in \mathcal{V}^{\mathcal{T}} \Big\}.
\end{align}
In words, for each target-domain item $x_i$ (except the last in the sequence), the decoder predicts the \emph{next occurring} target-domain item $x_t$, using the full mixed-domain history $\mathbf{x}_{1:i}$ as context. This approach allows the model to fully utilize cross-domain patterns learned by the encoder while enabling domain-specific decoders to develop the specialized expertise necessary for accurate target-domain predictions. The same SDP strategy is also implemented for every source domain during their respective adaptation stages.

\subsection{Adaptive Contrastive Decoding}

This process is designed to address two key challenges: (1) determining whether an item from a mixed-domain sequence should be converted into a target-domain item, and (2) if transformation is appropriate, deciding which target-domain item it should be transformed into. After obtaining the cross-domain decoder and the domain-specific decoder in the previous stage, we draw inspiration from the conventional "Expert-Amateur" contrastive decoding paradigm. In this setup, the target-domain expert $\theta^{*}_{\mathcal{T}}$ serves as a strong supervisor, providing high-quality domain knowledge. The source-domain experts function as Amateurs relative to the target domain, providing guidance on filtering potentially harmful cross-domain biases. Meanwhile, a third-party base model $\theta^{*}_{\mathcal{M}}$ is employed to quantify knowledge that is transferable across domains.

\subsubsection{Contrastive Score Calculation} 
When $x_{i+1}$ originates from the source domain $k$ within a mixed sequence, the historical sequence $\mathbf{x}_{1:i}$ is fed in parallel to the source-domain expert $\mathcal{P}^{\mathcal{S}_k}$, the target-domain expert $\mathcal{P}^{\mathcal{T}}$, and the base model $\mathcal{P}^{\mathcal{M}}$, yielding unnormalized prediction scores. To balance cross-domain transferability and target-domain specificity, we first compute the \emph{Global Contrastive Score} over the entire item space $\mathcal{V}$:
\begin{align}
    \textit{Global-score}_{i+1} = \alpha_g \cdot \mathcal{P}^{\mathcal{T}} - \beta_g \cdot \mathcal{P}^{\mathcal{S}_k}~,
\end{align}
where $\alpha_g$ and $\beta_g$ are the domain confidence and transfer reward coefficients, respectively. A source-domain item whose global score is maximized for a target-domain item indicates that the target-domain expert is highly confident about the transformation, and that the item also possesses transferable patterns across domains. Otherwise, the item is unlikely to require transformation.  For items deemed convertible, we further compute the \emph{Local Contrastive Score} restricted to the target-domain item set $\mathcal{V}^{\mathcal{T}}$:
\begin{align}
    \textit{Local-score}_{i+1} = \alpha_l \cdot \mathcal{P}^{\mathcal{T}}_{\mathcal{V}^{\mathcal{T}}} - \beta_l \cdot \mathcal{P}^{\mathcal{S}_k}_{\mathcal{V}^{\mathcal{T}}}~.
\end{align}
By focusing exclusively on target-domain candidates, the local score precisely identifies the most suitable item for transformation, effectively combining domain-specific confidence and cross-domain transferability. This two-level contrastive design enables the model to distill transferable patterns while retaining domain-specialized discriminative power.

\subsubsection{Adaptive Weight Decision} 
To enhance the adaptivity of knowledge integration, we define the domain confidence coefficient $\alpha$ and the transfer reward coefficient $\beta$ from an information-theoretic perspective, leveraging entropy and divergence principles. Specifically, $\alpha$ quantifies the informational certainty of the target-domain expert using the Shannon entropy of its output distribution. Lower entropy indicates a peaked, confident distribution, where the expert assigns high probability to specific items. Accordingly, positions with lower entropy should contribute more during decoding. To normalize $\alpha$ into $[0,1]$, we define:
\begin{equation}
\alpha_g = \frac{1 - H(\mathcal{P}^{\mathcal{T}}(v \mid \mathbf{x}_{1:i})))}{1 - \max_{v \in \mathcal{V}} H(\mathcal{P}^{\mathcal{T}}(v \mid \mathbf{x}_{1:i}))}.
\end{equation}
From this viewpoint, $\alpha$ serves as an information-certainty weight, amplifying the target expert’s influence in confident regions while reducing its effect in uncertain areas.

Meanwhile, $\beta$ captures the transferability of cross-domain knowledge via the Jensen–Shannon Divergence (JSD) between the base model and the source-domain expert. As a symmetrized and bounded measure, lower JSD indicates stronger alignment and hence higher transferability, whereas higher divergence highlights domain-specific conflicts. For the prediction distributions $\mathcal{P}^{\mathcal{M}}$ and $\mathcal{P}^{\mathcal{S}_k}$ at the current position, $\beta_g$ is defined as following:
\begin{equation}
\beta_g = \frac{1}{2} \Big[ \text{KL}\big(\mathcal{P}^{\mathcal{M}} \,\|\, \frac{\mathcal{P}^{\mathcal{M}} + \mathcal{P}^{\mathcal{S}_k}}{2} \big) 
+ \text{KL}\big(\mathcal{P}^{\mathcal{S}_k} \,\|\, \frac{\mathcal{P}^{\mathcal{M}} + \mathcal{P}^{\mathcal{S}_k}}{2} \big) \Big].
\end{equation}
Thus, $\beta$ serves as a transfer-alignment weight, reinforcing knowledge in regions where the base model and source expert are strongly aligned while attenuating contributions from positions with high cross-domain conflict. The same principles are applied locally to $\alpha_l$ and $\beta_l$, computed are restricted over the target-domain vocabulary.

\begin{comment}
\end{comment}

\vspace{8pt}
\section{Experiment}
    In this section, we empirically evaluate \textsc{Taesar} in terms of its effectiveness compared with mode-centric methods, generalizability across different architectures, and the rationality of its design.

\subsection{Experimental Settings}

\subsubsection{Datasets.}
    \def\arraystretch{1.3}
\setlength{\tabcolsep}{0.3em} % for the horizontal padding
\begin{table}[thbp]
    \caption{Statistics of experimental datasets.}
    % \begin{small}
    \begin{center}
        \begin{tabular}{c|c c c c c}
            \toprule
            \rowcolor[HTML]{F5F5F5} & \textbf{\#~user} & \textbf{\#~item} & \textbf{\#~inter} & \textbf{Avg.} & \\
            \rowcolor[HTML]{F5F5F5} \multirow{-2}{*}{\textbf{Dataset}} & (~$|\mathcal{U}|$~) & (~$|\mathcal{V}|$~) & (~$\sum_{\mathbf{x}}N_{\mathbf{x}}$~) & \textbf{length} & \multirow{-2}{*}{\textbf{sparsity(\%)}} \\
            \midrule
            Books       &  & 70,672 & 526,955 & 13.12 & 99.98 \\
            Electronics &  & 44,278 & 501,759 & 12.49 & 99.97 \\
            Sports      &  & 38,030 & 355,807 & 8.86 & 99.98 \\
            Tools       & \multirow{-4}{*}{40,165} & 31,880 & 331,230 & 8.25 & 99.97 \\
            \bottomrule
        \end{tabular}
    \end{center}
    % \end{small}
    \label{tab:stats}
\end{table}

    Publicly available Amazon review datasets were employed in our experiments\footnote{http://snap.stanford.edu/data/amazon/productGraph/categoryFiles/}. We randomly selected four domains: \emph{Books}, \emph{Electronics}, \emph{Sports and Outdoors}, and \emph{Tools and Home Improvement}. For clarity, these are referred to as Books, Electronics, Sports, and Tools, respectively and follow the preprocessing in~\citep{SyNCRec-park2024pacer, ABXI-bian2025abxi} to guarantee a minimum of 3 interactions associated with each user and item, then, retain only the 128 most recent interactions for longer sequences. Finally, user interactions are formatted chronologically. Table~\ref{tab:stats} summarizes the statistics of the resulting datasets.

\def\arraystretch{1.2}
\setlength{\tabcolsep}{4pt} % for the horizontal padding
\begin{table*}[htbp]
    \caption{Overall performance of \textsc{Taesar} compared with state-of-the-art multi-domain sequential recommendation methods, and its generalization across single-domain models with different architectures. Best results are in bold, second-best underlined.}
    \vspace{-10pt}
    \begin{small}
        \begin{center}
            \begin{tabular}{c|c c c c c c|c c c c c c}
                \toprule
                \multicolumn{13}{c}{\textbf{BEST: Books, Electronics, Sports, Tools}} \\
                \midrule
                \rowcolor[HTML]{E6E6E6} & \multicolumn{6}{c|}{\textbf{Books}} & \multicolumn{6}{c}{\textbf{Electronics}} \\
                \rowcolor[HTML]{E6E6E6} \multirow{-2}{*}{\textbf{Method}} & HR@10 & HR@20 & NG@10 & NG@20 & MRR@10 & MRR@20 & HR@10 & HR@20 & NG@10 & NG@20 & MRR@10 & MRR@20 \\
                \midrule
                SASRec            & 0.0563 & 0.0660 & 0.0371 & 0.0395 & 0.0311 & 0.0317 & 0.0350 & 0.0430 & 0.0257 & 0.0277 & 0.0228 & 0.0234 \\
\rowcolor[HTML]{F5F5F5} GRU4Rec   & 0.0384 & 0.0450 & 0.0283 & 0.0300 & 0.0253 & 0.0257 & 0.0247 & 0.0325 & 0.0184 & 0.0204 & 0.0165 & 0.0170 \\
                GCE-GNN           & 0.0598 & 0.0733 & 0.0447 & 0.0481 & 0.0400 & 0.0409 & 0.0353 & 0.0424 & 0.0286 & 0.0304 & 0.0266 & 0.0271 \\
\rowcolor[HTML]{F5F5F5} CL4SRec   & 0.0585 & 0.0708 & 0.0442 & 0.0473 & 0.0398 & 0.0406 & 0.0347 & 0.0424 & 0.0278 & 0.0298 & 0.0257 & 0.0262 \\
                \midrule
                $\text{SASRec}_4$ & 0.0504 & 0.0575 & 0.0342 & 0.0359 & 0.0291 & 0.0295 & 0.0311 & 0.0389 & 0.0240 & 0.0260 & 0.0218 & 0.0224 \\
% \rowcolor[HTML]{F5F5F5} C2DSR     & - & - & - & - & - & - & - & - & - & - & - & - \\
\rowcolor[HTML]{F5F5F5} ABXI      & 0.0489 & 0.0549 & 0.0341 & 0.0356 & 0.0369 & 0.0370 & 0.0352 & 0.0423 & 0.0290 & 0.0308 & 0.0246 & 0.0248 \\
                SyNCRec           & 0.0455 & 0.0564 & 0.0337 & 0.0367 & 0.0259 & 0.0262 & 0.0314 & 0.0402 & 0.0252 & 0.0273 & 0.0193 & 0.0197 \\
\rowcolor[HTML]{F5F5F5} CGRec     & 0.0413 & 0.0596 & 0.0316 & 0.0348 & 0.0218 & 0.0221 & 0.0338 & 0.0429 & 0.0203 & 0.0247 & 0.0190 & 0.0194 \\
                \midrule
                SASRec+\textsc{Taesar}            & 0.0613 & 0.0738 & 0.0406 & 0.0438 & 0.0342 & 0.0351 & \textbf{0.0441} & \textbf{0.0527} & 0.0291 & 0.0312 & 0.0243 & 0.0249 \\
\rowcolor[HTML]{F5F5F5} GRU4Rec + \textsc{Taesar} & 0.0407 & 0.0487 & 0.0307 & 0.0328 & 0.0277 & 0.0282 & 0.0288 & 0.0346 & 0.0233 & 0.0248 & 0.0216 & 0.0220 \\
                GCE-GNN+\textsc{Taesar}           & \textbf{0.0774} & \textbf{0.0945} & \ul{0.0566} & \textbf{0.0609} & \ul{0.0502} & \ul{0.0514} & 0.0404 & 0.0499 & \textbf{0.0319} & \textbf{0.0343} & \textbf{0.0293} & \textbf{0.0300} \\
\rowcolor[HTML]{F5F5F5} CL4SRec + \textsc{Taesar} & \ul{0.0772} & \ul{0.0912} & \textbf{0.0567} & \ul{0.0603} & \textbf{0.0504} & \textbf{0.0514} & \ul{0.0408} & \ul{0.0506} & \ul{0.0317} & \ul{0.0342} & \ul{0.0289} & \ul{0.0296} \\
                \bottomrule
            \end{tabular}

            \begin{tabular}{c|c c c c c c|c c c c c c}
                \toprule
                % \multicolumn{13}{c}{\textbf{BEST: Books, Electronics, Sports, Tools}} \\
                % \midrule
                \rowcolor[HTML]{E6E6E6} & \multicolumn{6}{c|}{\textbf{Sports}} & \multicolumn{6}{c}{\textbf{Tools}} \\
                \rowcolor[HTML]{E6E6E6} \multirow{-2}{*}{\textbf{Method}} & HR@10 & HR@20 & NG@10 & NG@20 & MRR@10 & MRR@20 & HR@10 & HR@20 & NG@10 & NG@20 & MRR@10 & MRR@20 \\
                \midrule
                SASRec            & 0.0399 & 0.0472 & 0.0295 & 0.0314 & 0.0262 & 0.0267 & 0.0333 & 0.0408 & 0.0249 & 0.0268 & 0.0222 & 0.0227 \\
\rowcolor[HTML]{F5F5F5} GRU4Rec   & 0.0262 & 0.0319 & 0.0202 & 0.0216 & 0.0183 & 0.0187 & 0.0237 & 0.0286 & 0.0188 & 0.0201 & 0.0173 & 0.0177 \\
                GCE-GNN           & 0.0379 & 0.0460 & 0.0318 & 0.0339 & 0.0300 & 0.0305 & 0.0332 & 0.0389 & 0.0270 & 0.0285 & 0.0251 & 0.0255 \\
\rowcolor[HTML]{F5F5F5} CL4SRec   & 0.0413 & 0.0471 & 0.0335 & 0.0350 & 0.0311 & 0.0315 & 0.0325 & 0.0391 & 0.0268 & 0.0285 & 0.0251 & 0.0255 \\
                \midrule
                $\text{SASRec}_4$ & 0.0373 & 0.0419 & 0.0281 & 0.0292 & 0.0251 & 0.0254 & 0.0319 & 0.0362 & 0.0236 & 0.0247 & 0.0210 & 0.0213 \\
% \rowcolor[HTML]{F5F5F5} C2DSR     & - & - & - & - & - & - & - & - & - & - & - & - \\
\rowcolor[HTML]{F5F5F5} ABXI      & 0.0386 & 0.0438 & 0.0306 & 0.0320 & 0.0254 & 0.0255 & 0.0309 & 0.0357 & 0.0271 & 0.0283 & 0.0232 & 0.0234 \\
                SyNCRec           & 0.0368 & 0.0395 & 0.0306 & 0.0343 & 0.0244 & 0.0247 & 0.0371 & 0.0431 & 0.0307 & 0.0329 & 0.0206 & 0.0230 \\
\rowcolor[HTML]{F5F5F5} CGRec     & 0.0344 & 0.0384 & 0.0299 & 0.0316 & 0.0224 & 0.0228 & 0.0344 & 0.0399 & 0.0202 & 0.0237 & 0.0215 & 0.0228 \\
                \midrule
                SASRec+\textsc{Taesar}            & 0.0568 & 0.0667 & 0.0354 & 0.0379 & 0.0287 & 0.0294 & 0.0465 & \textbf{0.0573} & 0.0288 & 0.0315 & 0.0232 & 0.0240 \\
\rowcolor[HTML]{F5F5F5} GRU4Rec + \textsc{Taesar} & 0.0322 & 0.0379 & 0.0252 & 0.0266 & 0.0231 & 0.0235 & 0.0267 & 0.0326 & 0.0217 & 0.0231 & 0.0201 & 0.0205 \\
                GCE-GNN+\textsc{Taesar}           & \ul{0.0582} & \ul{0.0676} & \ul{0.0445} & \ul{0.0469} & \ul{0.0403} & \ul{0.0409} & \ul{0.0482} & 0.0568 & \ul{0.0377} & \ul{0.0398} & \ul{0.0344} & \ul{0.0350} \\
\rowcolor[HTML]{F5F5F5} CL4SRec + \textsc{Taesar} & \textbf{0.0602} & \textbf{0.0698} & \textbf{0.0464} & \textbf{0.0488} & \textbf{0.0421} & \textbf{0.0428} & \textbf{0.0491} & \ul{0.0568} & \textbf{0.0385} & \textbf{0.0405} & \textbf{0.0352} & \textbf{0.0358} \\
                \bottomrule
            \end{tabular}

        \end{center}
    \end{small}
    \label{tab:overall}
    \vspace*{-1\baselineskip}
    % \vspace*{0.1\baselineskip}
\end{table*}

\subsubsection{Baselines and Target Models.}

    To fully demonstrate the cross-architecture generalizability of \textsc{Taesar}, we select target models from various model architecture categories:
    \begin{itemize}[topsep=4pt, itemsep=2pt, leftmargin=0.5cm]
        \item \textbf{\emph{RNN-Based:}} \emph{GRU4Rec}\citep{GRU4Rec-hidasi2015session} models user interaction sequences with item embeddings processed by GRUs and optimizes ranking-based objectives to improve next-item prediction accuracy.
        \item \textbf{\emph{GNN-Based:}} \emph{GCE-GNN}\citep{GCE-GNN-wang2020global} learns item embeddings from both the current session and all global sessions then uses GNN on each graph and combines them with a soft attention mechanism.
        \item \textbf{\emph{Attention-Based:}} \emph{SASRec}\citep{SASRec-kang2018self} models user sequences with self-attention to capture long-term dependencies.
        \item \textbf{\emph{Contrastive-Learning-Based:}} \emph{CL4SRec}\citep{cl4srec-xie2022contrastive} stands out as a powerful contrastive sequential recommender by introducing three effective unique sequence-level augmentation strategies.
    \end{itemize}
    To verify the \textsc{Taesar}'s superiority compared with cross-domain methods, we select the following representative baselines:
    \begin{itemize}[topsep=4pt, itemsep=2pt, leftmargin=0.5cm]
        \item \textbf{\emph{$\text{SASRec}_m$}} is a directly variant using SASRec~\citep{SASRec-kang2018self} trained on naive mixing data to revealing performance degradation.
        % \item \textbf{\emph{C2DSR}} instantiate difference set of graphical and self-attention encoder to encode cross-domain and domain-specific sequences, leveraging augmentation for contrastive learning.
        \item \textbf{\emph{ABXI}}\citep{ABXI-bian2025abxi} introduces a task-guided alignment mechanism to address the prediction mismatches and employs two types of LoRA to adaptively integrates domain-invariant interests.
        \item \textbf{\emph{CGRec}}\citep{CGRec-park2023cracking} reweights domain loss via hierarchical contrastive learning to capture cross-domain correlations.
        \item \textbf{\emph{SyNCRec}}\citep{SyNCRec-park2024pacer} uses single-domain and multi-domain contrastive learning to control negative transfer and an auxiliary mutual information loss to enhance representation transfer across tasks.
    \end{itemize}

\subsubsection{ Evaluation Protocols.}

    We rigorously evaluate the next-item prediction performance using the widely adopted leave-one-out split strategy. Specifically, for each sequence, the most recent interaction is reserved for testing, the second-to-last for validation, and the remainder for training. Consistent with standard cross-domain practices, we construct domain-specific validation and test sets based on the domain of the last item in each sequence. Performance is assessed using three established ranking metrics: Hit Rate ($\text{HR}@k$), Normalized Discounted Cumulative Gain ($\text{NG}@k$), and Mean Reciprocal Rank ($\text{MRR}@k$), with $k \in \{10, 20\}$. Moreover, to address the observation by Krichene~\citep{krichene2020sampled} that "sampled metrics" (metrics computed using sampled negative items) can yield results inconsistent with full metrics, meaning the relative performance ranking between different algorithms may not remain stable, we adopt the entire item set as the candidate item set during evaluation.

\subsubsection{Implementation Details.}

    We implemented \textsc{Taesar} using PyTorch and developed single-domain sequential recommendation models based on the Recbole library. For the multi-domain baselines, we utilized their publicly available source code. Crucially, we replaced the sampled metrics with full metrics to ensure alignment with our evaluation methodology, as discussed previously. Notably, although ABXI was initially demonstrated in a two-domain setting, the original work established its straightforward scalability to multiple domains. Consequently, we adapted and extended both methods to effectively operate in our multi-domain environment.
    
    To ensure a fair comparison, all single-domain models were uniformly configured across all datasets (including the \textsc{Taesar}-regenerated dataset) with the following fixed hyperparameters: $2$ attention heads and layers, hidden and inner sizes of $128$, and dropout rates of $0.2$. Conversely, the multi-domain models were subjected to hyperparameter search over the following space: attention heads $\in \{1, 2, 3\}$, layers $\in \{1, 2, 3\}$, hidden size $\in \{64, 128, 256\}$, and inner size $\in \{64, 128, 256\}$. Specifically for ABXI, the LoRA rank $\in \{8, 16, 32\}$, and for SyNCRec, the number of experts $\in \{4, 6, 8\}$. All models utilize the Adam optimizer with a learning rate $\in \{10^{-3}, 3 \times 10^{-4}, 10^{-4}\}$. The training process runs for a maximum of $300$ epochs, and early stopping is employed if the $\text{NG}@10$ on the validation data fails to improve for $30$ consecutive epochs.

\subsection{Overall Performance}

\subsubsection{How superior is \textsc{Taesar} to model-centric methods?}

Experimental results reveal that cross-domain recommendation performance is highly sensitive to data processing strategies. As shown in Table~\ref{tab:overall}:
(1) \textbf{\emph{Naïve multi-domain merging leads to negative transfer.}}
The performance of SASRec4 falls short of the single-domain baseline SASRec, indicating that simply merging data from multiple domains—without any adaptation—induces severe negative transfer due to distributional heterogeneity, thereby compromising the target domain’s performance.
(2) \textbf{\emph{Model-centric improvements are limited under weak inter-domain semantics.}}
Traditional model-centric approaches occasionally outperform both SASRec4 and single-domain baselines on specific metrics; however, their overall performance tends to degrade. This suggests that when semantic correlations across domains are weak, architectural enhancements alone cannot effectively bridge inter-domain discrepancies.
(3) \textbf{\emph{\textsc{Taesar} effectively mitigates inter-domain interference.}}
In contrast, our data-centric framework \textsc{Taesar} achieves consistent and significant improvements over both SASRec4 and single-domain models. This demonstrates that \textsc{Taesar} alleviates inter-domain interference through data-level refinement, enabling precise and domain-specific modeling without being hindered by heterogeneous knowledge.
Collectively, these findings underscore that as model-level optimizations approach diminishing returns, structured data regeneration—the core of \textsc{Taesar}—offers a more robust and interpretable solution for cross-domain recommendation.

\def\arraystretch{1.4}
\setlength{\tabcolsep}{2pt}
\begin{table}[!t]
    \caption{Ablation results of \textsc{Taesar} on different designs of Pretraining, Domain-Specific Adaptation, and Adaptive Contrastive Decoding strategies.}
    \begin{small}
        \begin{center}
        \begin{tabular}{l | c c c c c c}
            \toprule
            \multicolumn{7}{c}{\textbf{Electronics}} \\
            \midrule
            \rowcolor[HTML]{E6E6E6} \textbf{Variants} & HR@10 & HR@20 & NG@10 & NG@20 & MRR@10 & MRR@20 \\
            \midrule
            w/o DSA & 0.0352 & 0.0434 & 0.0237 & 0.0258 & 0.0202 & 0.0207 \\
            DSA w/o DSP & 0.0350 & 0.0450 & 0.0241 & 0.0266 & 0.0208 & 0.0215 \\
            \midrule
            DSP w/o SDE & 0.0398 & 0.0472 & 0.0282 & 0.0300 & 0.0239 & 0.0247 \\
            DSP w/o GCS & 0.0439 & 0.0523 & 0.0289 & 0.0310 & 0.0242 & 0.0247 \\
            DSP w/o LCS & 0.0431 & 0.0512 & 0.0278 & 0.0299 & 0.0231 & 0.0236 \\
            \midrule
            \textsc{Taesar} Full & \textbf{0.0441} & \textbf{0.0527} & \textbf{0.0291} & \textbf{0.0312} & \textbf{0.0243} & \textbf{0.0249} \\
            \bottomrule
        \end{tabular}
        \end{center}
    \end{small}
    \label{tab:ablation}
    \vspace*{-1\baselineskip}
\end{table}

\subsubsection{How versatile is regenerated data from \textsc{Taesar}?}

Although \textsc{Taesar} employs a decoder-based architecture to perform contrastive data regeneration, the regenerated data itself remains model-agnostic. To assess its generality, we train multiple models with distinct architectural designs on the regenerated dataset. As shown in Table~\ref{tab:overall},
(1) \textbf{\emph{Regenerated data generalizes across architectures.}}
Models with diverse architectures trained on \textsc{Taesar}’s regenerated data exhibit consistent and substantial performance improvements, confirming the model-agnostic generalization capability of the regenerated dataset.
(2) \textbf{\emph{More expressive architectures yield greater improvements.}}
When trained on the regenerated data, models with stronger representational capacity exhibit more pronounced performance gains. This observation suggests that the high-quality, semantically consistent data produced by \textsc{Taesar} allows expressive architectures to more effectively leverage their modeling capacity, thereby amplifying their advantage in capturing subtle user–item interactions.
(3) \textbf{\emph{Data-centric regeneration complements model-centric design.}}
The consistent cross-architecture gains underscore the inherent complementarity between paradigms: model-centric approaches enhance structural expressiveness, whereas \textsc{Taesar}’s data-centric paradigm fundamentally strengthens the learning signal, providing a more robust basis for future advancing recommender systems.

\subsection{Further Analysis}

\subsubsection{How indispensable is the architecture of \taesar?}

Table~\ref{tab:ablation} presents a comprehensive ablation study evaluating the design choices of \textsc{Taesar}. 
For the first stage, \emph{Pretraining and Domain-Specific Adaptation}, we investigate the necessity of \emph{domain-specific adaptation} (DSA) and the effectiveness of \emph{domain-specific prediction} (DSP) within DSA. Specifically, `w/o DSA' refers to training the domain-specific experts directly using DSP without adapting from the base model, whereas `DSA w/o DSP' denotes adapting from the base model using only single-domain sequence data without employing DSP. The results indicate that DSA coupled with DSP consistently yields the best performance, highlighting the critical role of this combination.
For the second stage, \emph{Adaptive Contrastive Decoding}, we examine the contributions of \emph{source-domain experts} (Amateur) and the two-level contrastive design. Here, `w/o SDE' removes the source-domain experts during decoding, relying solely on the target-domain expert’s prediction distribution for cross-domain item transformation. `w/o GCS' and `w/o LCS' correspond to using only the local or global contrastive scores, respectively. Across all these variants, performance consistently drops compared to the full \textsc{Taesar} model, demonstrating the importance of both the source-domain knowledge and the two-level contrastive mechanism.
Overall, these ablation results clearly illustrate that each architectural component of \textsc{Taesar}—from DSA with DSP to adaptive contrastive decoding—is essential for achieving optimal cross-domain sequential recommendation performance.

\subsubsection{Is domain-shared information successfully detected and effectively transferred?}

\begin{figure}
\centering
\includegraphics[width=.85\linewidth]{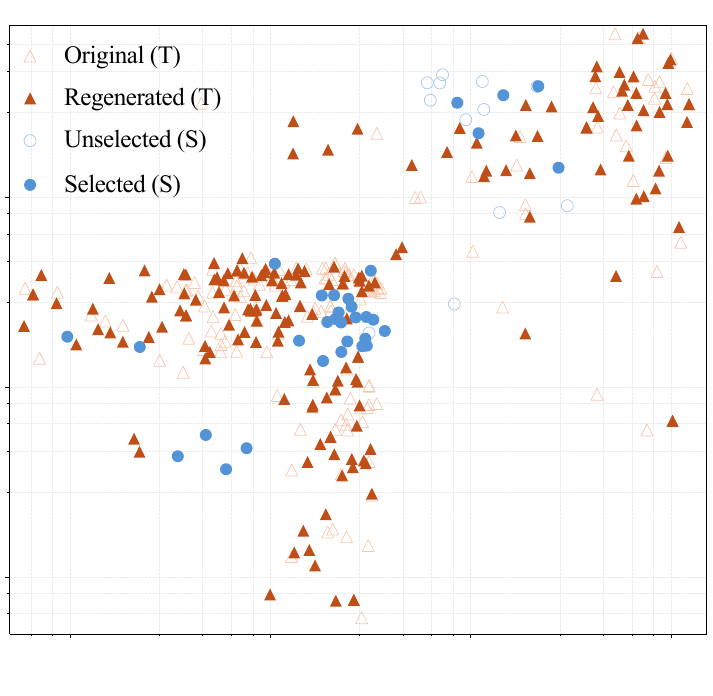}
\vspace{-10pt}
\caption{Schematic illustration of the adaptive contrastive decoding phase: (a) selection of transferable items from source domain $\mathcal{S}$ (Electronics), and (b) distribution-level mapping to target domain $\mathcal{T}$ (Books).}
\label{fig:tsne}
\end{figure}

To assess whether \textsc{Taesar} effectively identifies and transfers domain-shared information, we analyze the item-level mappings produced during adaptive contrastive decoding. Specifically, we extract the mapping dictionary from source items to their corresponding target items and visualize the learned item embeddings from both domains. As shown in Figure~\ref{fig:tsne}, blue points represent source-domain items, with solid blue dots indicating items selected for transformation into target-domain counterparts, corresponding to solid red dots. The visualization reveals that these selected source items have embeddings closely aligned with those of the target items, suggesting that they capture transferable, domain-shared semantics. This alignment demonstrates that \textsc{Taesar} successfully identifies and transforms transferable items, highlighting its effectiveness in utilizing cross-domain knowledge.

\begin{figure}[thbp]
    \centering
    \begin{subfigure}[b]{0.5\columnwidth}
        \centering
        \includegraphics[width=0.90\textwidth]{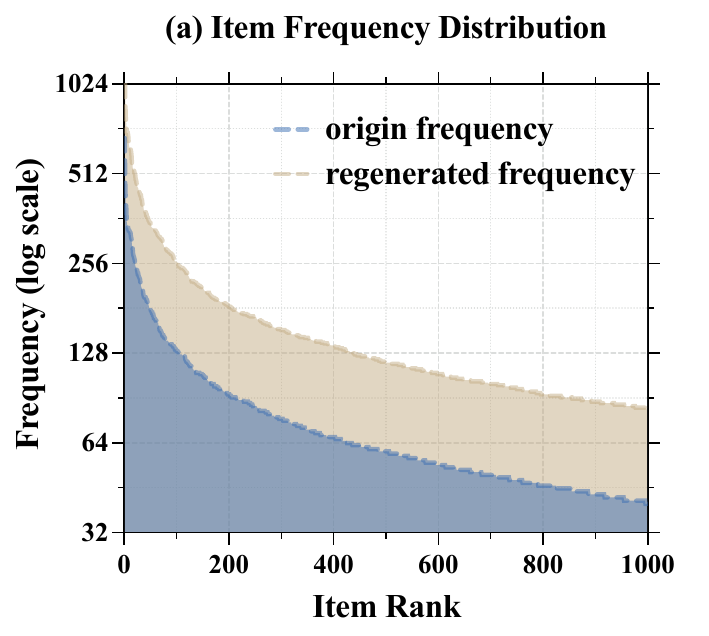}
        \label{fig:a}
    \end{subfigure}
    % \hfill
    \hspace{-3pt}
    \begin{subfigure}[b]{0.495\columnwidth}
        \centering
        \includegraphics[width=0.90\textwidth]{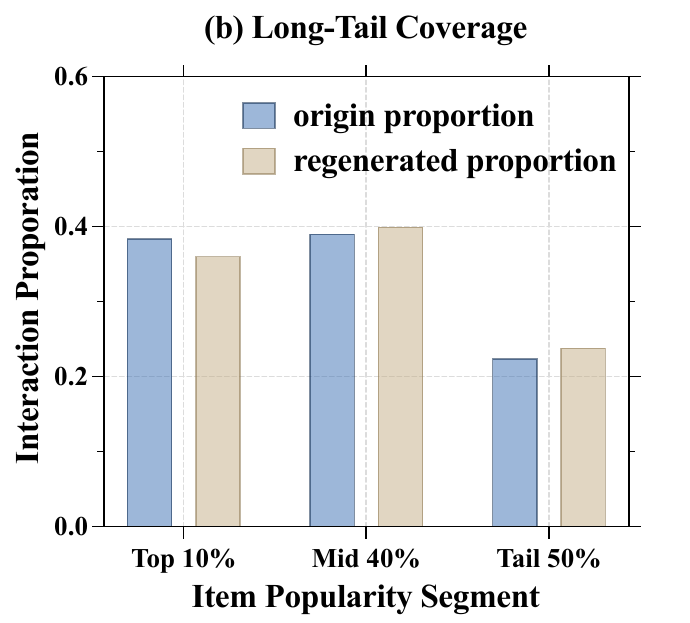}
        \label{fig:b}
    \end{subfigure}

    % \vspace{-1em}
    \begin{subfigure}[b]{1\columnwidth}
        \centering
        \includegraphics[width=0.90\textwidth]{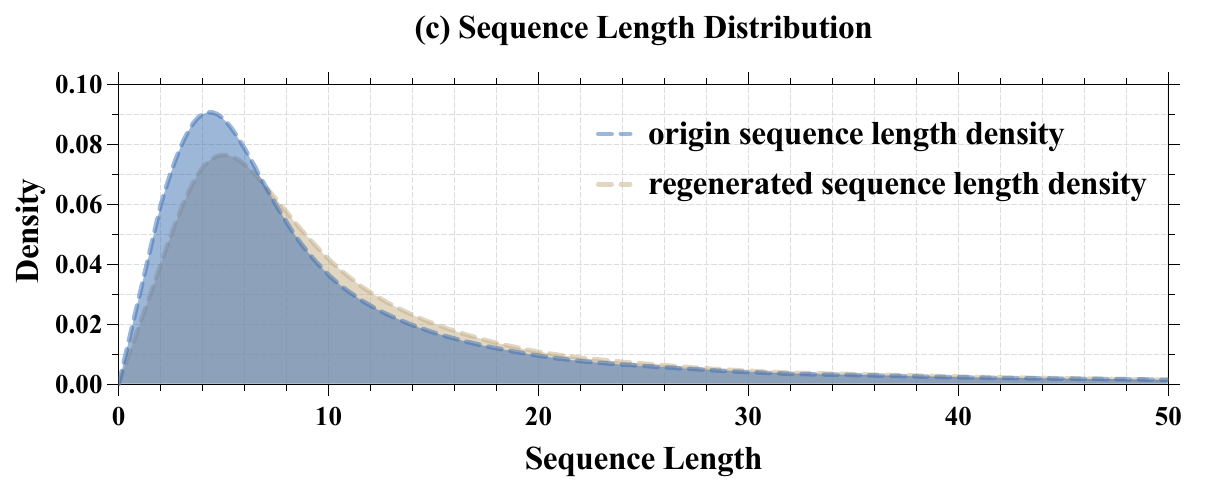}
        \label{fig:c}
    \end{subfigure}

    \vspace{-1em}
    \caption{Comparison of data statistics between the original and regenerated Books domain dataset: (a) item frequency, (b) long-tail coverage, and (c) sequence length distribution.}
    
    \label{fig:dataset_comparison}
\end{figure}

\subsubsection{How regenerated dataset enhances diversity and context?} Analysis of the regenerated dataset in the Books domain reveals that our data-centric regeneration substantially improves target-domain distributions and sequence context, explaining the observed performance gains. Compared with the original dataset, regenerated dataset exhibits a flatter item frequency curve (Figure~\ref{fig:dataset_comparison}~(a)), activating more low-frequency items and mitigating head dominance. Long-tail coverage is also enhanced (Figure~\ref{fig:dataset_comparison}~(b)), with reduced Top 10\% and increased Mid 40\% and Tail 50\% proportions, indicating the injection of diverse, previously underrepresented interactions. Moreover, user sequences in the regenerated dataset are longer and denser (Figure~\ref{fig:dataset_comparison}~(c)), providing richer temporal context and enabling models to capture extended sequential patterns. Collectively, these improvements in item-level diversity and sequence-level context highlight how data-centric regeneration enables robust and interpretable performance gains.

\vspace{8pt}

\section{Conclusion}

    \vspace{5pt}
    
    In this work, we present \textsc{Taesar}, a novel data-centric framework for cross-domain sequential recommendation. By prioritizing data transformation over complex transfer architecture designs, \textsc{Taesar} effectively mitigates negative transfer through target-aligned sequence regeneration. Specifically, it leverages adaptive contrastive decoding to identify transferable items from source domains and transform them into semantically aligned target-domain items, preserving temporal dynamics while eliminating potentially detrimental inter-domain information. Extensive experiments across multiple datasets and model architectures demonstrate that \textsc{Taesar} not only attains SOTA performance in cross-domain recommendation but also generalizes seamlessly to existing single-domain sequential recommenders without requiring modifications to their architectures. These findings highlight the synergy between data-centric and model-centric paradigms, providing a practical, theoretically grounded framework for cross-domain knowledge transfer.

    Despite these advances, important limitation remain. A deeper theoretical understanding of the limits of information preservation during the regeneration process is needed. Future work will focus on addressing above challenge, advancing both the theoretical foundations and practical applicability of data-centric recommendation.

% \input{tex/acknowledge}

%%
%% The next two lines define the bibliography style to be used, and
%% the bibliography file.

\bibliographystyle{ACM-Reference-Format}
\newpage
\bibliography{reference}

%%
%% If your work has an appendix, this is the place to put it.
% \appendix

\end{document}